\documentclass[conference]{IEEEtran}
\IEEEoverridecommandlockouts
\usepackage{cite}
\usepackage{amsmath,amssymb,amsfonts}
\usepackage{algorithmic}
\usepackage{graphicx}
\usepackage{textcomp}
\usepackage{xcolor}
\usepackage{polyglossia}
\setdefaultlanguage{english}
\setotherlanguage{arabic}

\newfontfamily\arabicfont[
    Script=Arabic,
    Path=./
]{Amiri-Regular.ttf}
\def\BibTeX{{\rm B\kern-.05em{\sc i\kern-.025em b}\kern-.08em
    T\kern-.1667em\lower.7ex\hbox{E}\kern-.125emX}}
\begin{document}

\title{Severity-Aware Curriculum Learning with Multi-Model Response Selection for Medical Text Generation\\
}

\author{\IEEEauthorblockN{Ahmed Alansary}
\IEEEauthorblockA{\textit{Faculty of Computer Science} \\
\textit{MSA University}\\
Giza, Egypt \\
ahmed.mohamed406@msa.edu.eg}
\and
\IEEEauthorblockN{Molham Mohamed}
\IEEEauthorblockA{\textit{Faculty of Computer Science} \\
\textit{MSA University}\\
Giza, Egypt \\
molham.mohamed@msa.edu.eg}
\and
\IEEEauthorblockN{Ali Hamdi}
\IEEEauthorblockA{\textit{Faculty of Computer Science} \\
\textit{MSA University}\\
Giza, Egypt \\
ahamdi@msa.edu.eg}}

\IEEEpubid{\makebox[\columnwidth]{ 979-8-3315-8488-7/26/\$31.00 ©2026 IEEE \hfill}
\hspace{\columnsep}\makebox[\columnwidth]{ }}
\maketitle
\IEEEpubidadjcol

\begin{abstract}
Telehealth systems have become increasingly important for delivering accessible and timely medical information. Existing large language models often struggle to provide consistent and contextually appropriate medical responses across varying levels of case severity. This limitation highlights the need for models that can effectively adapt to the progressive complexity in medical queries. To address this challenge, we introduce a severity-aware multi-model framework that integrates curriculum training strategy with relevance-based response selection. The proposed framework employs a three-stage curriculum learning strategy, where each model is trained sequentially on mild, moderate, and critical cases to progressively acquire domain knowledge. The approach uses five large language models, each trained independently under the same curriculum. During inference, all models generate candidate responses, and the response with highest BERTScore is selected as the final output. The framework is trained and evaluated on the MAQA dataset, which provides annotated medical question-answer pairs. Experimental results evaluated using BERTScore demonstrate that the proposed method achieves superior performance compared to both baseline and fine-tuned models, attaining 86.71\% in the baseline setting and 90.30\% after fine-tuning. These results highlight the effectiveness of combining curriculum learning with multi-model response selection in improving response quality and relevance in medical text generation.
\end{abstract}


\begin{figure*}
    \centering
    \renewcommand{\figurename}{Fig.}
\includegraphics[width=0.6\linewidth,height=5cm]{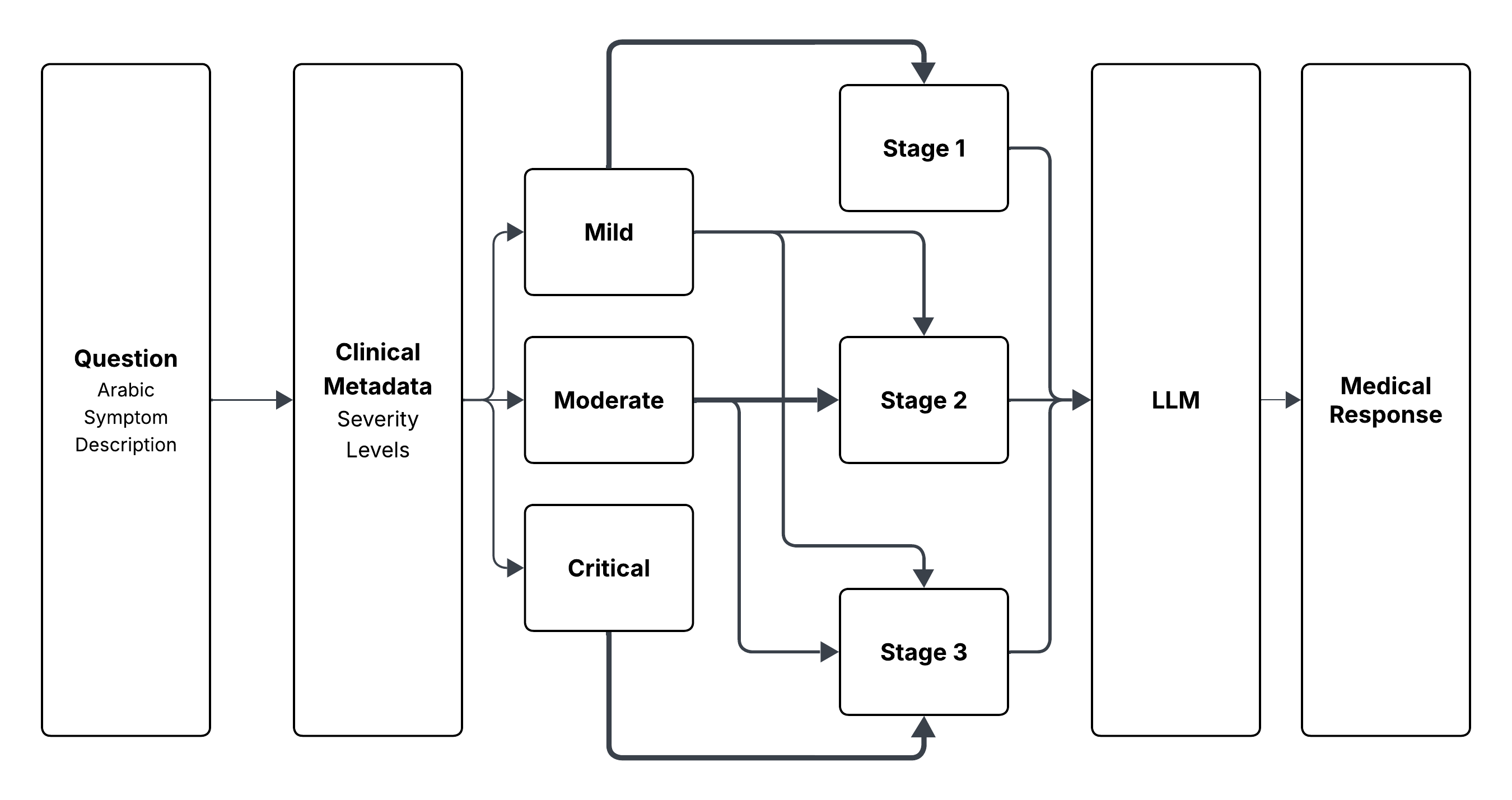}
    \caption{ Overview of the proposed severity-based curriculum learning framework}
    \label{fig:curriculum}
\end{figure*}

\section{Introduction}
Recent progress in telehealth infrastructure has reshaped how patients interact with healthcare providers, allowing remote and timely access to medical consultations\cite{p24}. Online healthcare platforms enable users to describe symptoms and obtain medical recommendations, producing large collections of health-related textual data that can support intelligent decision-making systems \cite{p4,p6}. In parallel, natural language processing (NLP) plays a key role in building intelligent systems for clinical support, facilitating the understanding and generation of medical text across various applications, including clinical support, personal care, and public health \cite{paper9,paper5}. The growing use of medical bots highlights the importance of generating responses that are both relevant and dependable in real-world healthcare settings. \cite{paper10,paper23,paper25}.

Despite these advances, existing approaches have several limitations. Recent progress in data-to-text generation has largely relied on neural end-to-end systems, where curriculum learning has been shown to improve both performance and speed by organizing training samples based on difficulty \cite{paper1}. Similarly, in-sample curriculum strategies that follow an easy-to-hard learning paradigm have demonstrated strong generalization capabilities in natural language generation tasks \cite{paper2}. In the context of healthcare applications, transformer-based architectures have shown competitive results across a range of clinical tasks including symptom analysis and disease categorization, symptom severity assessment, and medical text generation \cite{p4,p6,paper7}. Ensemble learning techniques improve prediction reliability by combining outputs from multiple models \cite{paper5}. Nevertheless, these approaches focus on classification or single-model generation, without considering how medical cases become more complex or how the quality of responses can change. Moreover, recent studies highlight challenges in evaluating LLM-generated responses in sensitive domains such as mental health, where reliability, safety, and relevance remain inconsistent \cite{paper3,paper22}. Additional challenges remain in Arabic medical NLP, including limited annotated resources, linguistic variability, and difficulties in handling diverse user inputs \cite{paper8,paper10}. Collectively, these limitations indicate the need for frameworks that can both adapt to varying case severity and enhance the reliability of generated medical responses.

To address these challenges, we introduce a severity-aware multi-model framework for medical text generation that combines curriculum training strategy with relevance-based response selection. The proposed approach employs a three-stage curriculum learning strategy, where models are trained sequentially on mild, moderate, and critical cases as illustrated in Figure \ref{fig:curriculum} to capture the progressive complexity of medical scenarios\cite{paper20}. Five large language models are trained independently under this curriculum scheme, ensuring diversity in learned representations. During inference, each model generates a candidate response to the user query, and a relevance-based scoring mechanism is used to select the most appropriate response as the final output. The framework is trained and evaluated on the MAQA dataset, a large-scale Arabic healthcare question-answer corpus \cite{paper10}. By combining curriculum learning with multi-model response selection, the proposed method aims to improve both the quality and reliability of medical text generation in telehealth applications.

\begin{table*}[t]
\centering
\caption{Example instances from the Arabic medical severity-aware dataset with transliteration and English translations.}
\label{tab:dataset_examples}
\scriptsize
\begin{tabular}{p{0.42\textwidth} p{0.40\textwidth} p{0.10\textwidth}}
\hline
\textbf{Question} & \textbf{Answer} & \textbf{Severity}\\
\hline

\textarabic{ورم في ارقبه كيف اتعامل معه هل يستدعي جراحه} &
\textarabic{علي حسب مكانه ونوعه روح لطبيب اورام} &
\textarabic{حرج} \\

Waram fi el raqaba keef atamel maah hal yastadey geraaha &
Ala hasab makano w noao rooh le tabeeb awram &
Harej \\

A neck lump, how should I deal with it? Does it require surgery? &
It depends on its location and type; consult an oncologist. &
Critical \\

\hline

\textarabic{اعاني من انتفاخ الخد نتيجه التورم اللثه بسبب تسوس الاسنان الماميه} &
\textarabic{يحتاج الامر لعلاج بمضاد حيوي وعمل علاج عصب} &
\textarabic{غير حرج} \\

Aani men entefakh el khad netiget el torom el latha besabab tasawos el asnan el amamiya &
Yahtag el amr le elag bedod hayawi wa amal elag asab &
Gher harej \\

I suffer from cheek swelling due to gum inflammation caused by front tooth decay. &
The condition requires antibiotic treatment and a root canal procedure. &
Mild \\

\hline

\textarabic{دايما احس بالم في الصدر من الجهه اليسار فوق الثدي} &
\textarabic{امرض القلب لاتعاطي الم مستمر ودائم احتمال يكون عندك شد عضلي اي تقلص في عضلات الصدر راجعي اي طبيب للفحص عليك} &
\textarabic{متوسط} \\

Dayman ahes be alam fi el sadr men el geha el yasar fow el thadi &
Amrad el qalb la taati alam mostamer wa daem ehtemal ykoon andak shad adali ay taqalos fi adalat el sadr rajei ay tabeeb lel fahs &
Motawaset \\

I always feel pain in the left side of my chest above the breast. &
Heart diseases do not usually cause constant pain; it may be a muscular strain. Consult a doctor for examination. &
Moderate \\

\hline
\end{tabular}
\end{table*}

\section{Related Work}

Recent research in healthcare language technologies have been significantly enhanced by transformer-based models and large-scale pretrained language architectures. Such models have shown promising performance in several medical tasks, including disease prediction, symptom analysis, clinical text understanding, and automated question answering \cite{p4,p6,paper7}. In the context of Arabic healthcare, several studies have highlighted the effectiveness of fine-tuning transformer models and leveraging domain-specific preprocessing techniques to improve performance, particularly when handling linguistically diverse and morphologically rich data \cite{p4,paper8}. Moreover, the availability of large-scale datasets such as MAQA has facilitated the development and evaluation of medical question-answering and text-generation systems, while also addressing challenges related to data scarcity in low-resource settings \cite{paper10,paper27}. Comprehensive surveys further emphasize the growing role of NLP and LLMs in smart healthcare, while identifying persistent challenges such as data limitations, domain-specific terminology, and evaluation reliability \cite{paper9,paper11,paper12,paper18}.

To reduce model-specific limitations, researchers have increasingly explored ensemble and multi-model solutions within biomedical NLP applications. Combining outputs from multiple systems has been reported to improve prediction consistency and overall performance across a variety of healthcare-related tasks \cite{paper5,p6}. Similarly, recent work in medical text generation has focused on fine-tuning multiple LLMs to produce coherent and contextually relevant responses, showing how generative AI can be used in healthcare applications \cite{paper7,paper21}. In parallel, synthetic data generation and knowledge integration techniques have been proposed to address data scarcity and improve performance \cite{paper14}. Despite these advances, most current methods still use either single-model generation or simple combining mechanisms, without explicitly modeling the variability in response quality or use query-aware selection strategies.

Curriculum learning has emerged as an effective paradigm for improving training efficiency and model performance by organizing training samples according to difficulty. Early work in data-to-text generation demonstrated that competence-based curriculum strategies can accelerate grouping and improve generation quality by dynamically selecting training samples based on model ability \cite{paper1}. After approaches extended this idea to natural language generation by adopting easy-to-hard learning schemes, including in-sample curriculum strategies that progressively increase generation complexity \cite{paper2}. Curriculum learning has also been applied in medical and multi-modal settings, where models are trained to handle increasingly complex clinical data, leading to improved generalization and performance across tasks \cite{paper13,paper17}. Additional studies have combined curriculum learning into diverse frameworks, such as dual learning for response generation and fine-tuning strategies for translation models, showing how flexible it is \cite{paper15,paper16}.

In the medical domain, recent efforts have focused on incorporating clinical relevance and case complexity into the training process\cite{paper26}. Severity-aware learning strategies have been proposed to prioritize critical medical cases, either through weighted loss functions or structured training schemes, leading to consistent performance improvements across multiple models \cite{paper19}. In particular, severity-based curriculum learning approaches organize training data into ordered stages (e.g., mild, moderate, and critical), allowing models to gradually adapt to increasing levels of complexity and better capture high-risk scenarios. Such strategies have shown promising results in Arabic medical text generation, improving response quality and robustness compared to conventional fine-tuning methods\cite{paper20}.

Despite these advancements, challenges remain in ensuring the reliability and evaluation of LLM-generated medical responses. Recent studies highlight that existing evaluation frameworks may lack consistency, particularly in sensitive domains such as mental health, where aspects such as safety, empathy, and relevance are difficult to assess reliably \cite{paper3}. These limitations underscore the need for approaches that not only improve training through structured learning strategies but also enhance inference by selecting the most appropriate response among multiple candidates.

\section{Dataset: Severity Annotation}
The \textit{Medical Arabic Question Answering (MAQA)} dataset was used in this work, which contains Arabic medical questions paired with their corresponding answers. The dataset was constructed for open-domain medical question answering and consists of textual pairs where each \textit{question} outlines a set of symptoms and each \textit{answer} delivers medical advice or guidance. For this study, a subset of 32K question-response pairs is utilized for both training and evaluation.

As the dataset does not originally include annotations indicating the urgency of medical conditions, an additional severity labeling scheme is introduced to facilitate the \textit{severity-based curriculum learning strategy}. In particular, each question is categorized into one of three severity levels: \textit{Mild}, \textit{Moderate}, or \textit{Critical}, as shown in Table~\ref{tab:dataset_examples}. The resulting labels reflect the relative severity of the reported symptoms and enable the creation of training subsets with increasing complexity.

The annotation procedure follows a rule-based methodology grounded in a curated list of Arabic medical symptom keywords. Initially, all questions undergo basic Arabic text normalization, including removing diacritics and punctuation and standardizing common letter variations. Subsequently, each question is examined for symptom-related keywords associated with varying levels of medical severity. Questions containing indicators of emergency conditions—such as severe chest pain, loss of consciousness, heavy bleeding, or difficulty breathing—are classified as \textit{Critical}. Those reflecting intermediate conditions, including fever, vomiting, or ongoing pain, are labeled as \textit{Moderate}. Questions describing less severe symptoms, such as headache, common cold, or minor discomfort, are assigned the \textit{Mild} label.

In cases where multiple keywords are present within a single question, the final label is determined by selecting the highest severity level based on the following priority: \textit{Critical} $>$ \textit{Moderate} $>$ \textit{Mild}. Questions that do not match any predefined symptom keywords are, by default, labeled as \textit{Mild}.

These severity annotations make it possible to partition the dataset into staged subsets that reflect increasing levels of medical complexity. This organization aligns with the proposed curriculum learning framework, whereby the model is first trained on mild cases and is then progressively exposed to moderate and critical cases during the fine-tuning phase.

\section{Methodology}

This proposed framework integrates severity-based curriculum learning with relevance-driven response selection for medical text generation. The methodology comprises two primary stages: curriculum-based training of multiple large language models (LLMs) and inference-time response selection according to semantic relevance.

\subsection{Curriculum-Based Training}

To enable the models to capture the progressive complexity of medical cases effectively, we adopt a severity-based curriculum learning strategy. Following the approach described in the recent work \cite{paper20}, the training data are partitioned into three severity-level subsets: mild, moderate, and critical.

\begin{equation}
    D_{\text{Mild}},\; D_{\text{Moderate}},\; D_{\text{Critical}}
\end{equation}

The training process is structured sequentially, with each model first trained on mild cases, then progressively exposed to moderate and critical cases.

The curriculum-based training process for each model \(M_m\), where \(m = 1,2,\dots,5\), is defined sequentially as follows:

\begin{equation}
    \theta_m^{(1)} = \operatorname{Train}(M_m, D_{\text{Mild}})
\end{equation}

\begin{equation}
    \theta_m^{(2)} = \operatorname{Train}(\theta_m^{(1)}, D_{\text{Moderate}})
\end{equation}

\begin{equation}
    \theta_m^{(3)} = \operatorname{Train}(\theta_m^{(2)}, D_{\text{Critical}})
\end{equation}

where \(\theta_m^{(3)}\) represents the final fine-tuned parameters of the \(m\)-th model after curriculum-based learning.

Five LLMs are independently fine-tuned using the same curriculum strategy. This staged training allows each model to first learn fundamental medical patterns from simpler cases before adapting to more complex and high-risk scenarios. By gradually increasing the difficulty of the training data, the models achieve improved generalization and stability compared to conventional training approaches.

All LLMs were fine-tuned using the Low-Rank Adaptation (LoRA) technique. The LoRA configuration was defined with rank \(r=16\), scaling factor \(\alpha=32\), and dropout rate \(0.05\). 
Each curriculum stage was trained for three epochs. To facilitate stable adaptation as the complexity of the medical cases increased, different learning rates were employed across the curriculum stages. Specifically, the learning rates were set to \(2e-5\), \(1e-5\), and \(8e-6\) for the first, second, and third stages, respectively.

\subsection{Response Generation}

After completing the training phase, each of the five LLMs is used to generate a response to a given patient query. Given an input question $q$, each model $M_i$ produces a candidate response $r_i$ according to

\begin{equation}
    r_i = M_i(q), \quad i = 1,2,\dots,5
\end{equation}

resulting in a set of generated responses:

\begin{equation}
    \{r_1, r_2, r_3, r_4, r_5\}.
\end{equation}

This multi-model generation process introduces diversity in the produced responses, as each model may capture different aspects of the input query and the underlying medical knowledge.

\begin{figure}[h]
    \centering
    \renewcommand{\figurename}{Fig.}
    \includegraphics[width=0.7\linewidth, height=6.2cm]{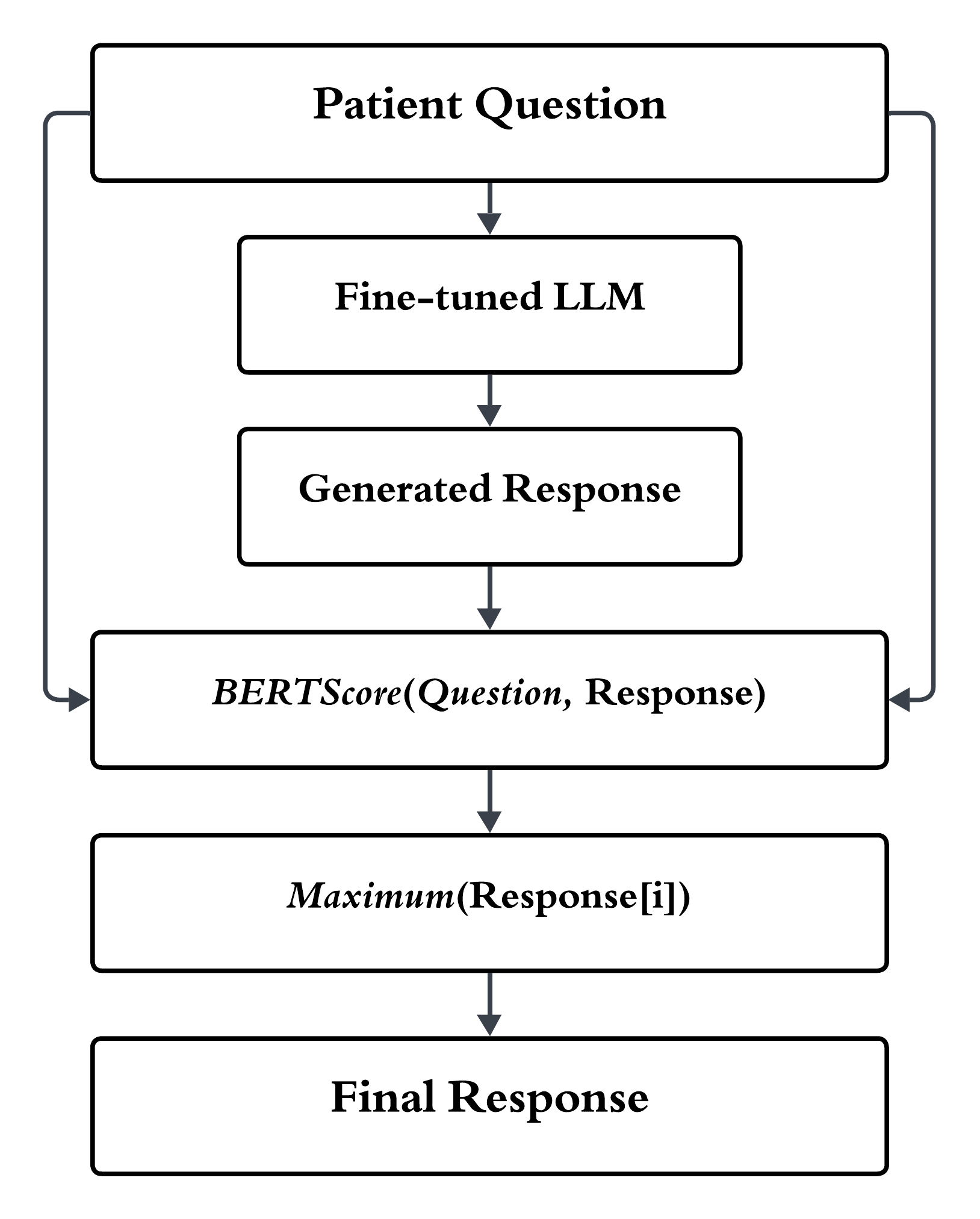}
    \caption{Overview of the proposed multi-model response selection.}
    \label{fig:multimodel}
\end{figure}

\subsection{Relevance-Based Response Selection}

To select the most appropriate response, we employ a relevance-based scoring mechanism using BERTScore. As illustrated in Figure \ref{fig:multimodel}, each generated response $r_i$ is compared with the input question $q$ to compute a semantic similarity score:

\begin{equation}
    s_i = \text{BERTScore}(r_i, q).
\end{equation}

The response with the highest similarity score is selected as the final output:

\begin{equation}
    r^* = \arg\max_{r_i} \; s_i.
\end{equation}

This selection strategy ensures that the chosen response is the most semantically aligned with the patient’s query, thereby improving the relevance and quality of the final output.

\subsection{Overall Framework}

The proposed methodology combines curriculum learning during training with a selection mechanism during inference. While curriculum learning enhances each individual model's ability to handle varying levels of medical complexity, the relevance-based selection step ensures that the most suitable response is delivered to the user. This combination provides a robust framework for medical text generation, particularly in telehealth scenarios where accuracy and contextual relevance are critical.

\section{Results}

\begin{figure*}
    \centering
    \renewcommand{\figurename}{Fig.}
    \includegraphics[width=0.7\linewidth, height=5cm]{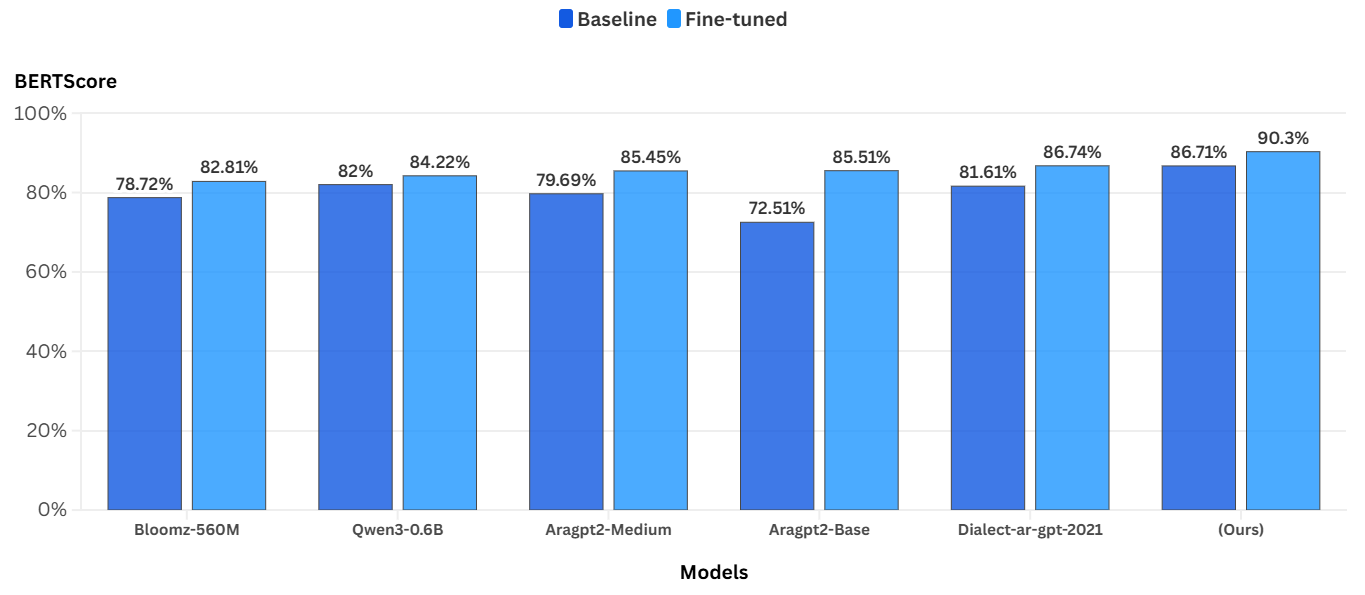}
    \caption{Visual comparison between baseline, fine-tuned models and Our proposed method.}
    \label{fig:Results_chart}
\end{figure*}

Table~\ref{tab:results_baseline} and Table~\ref{tab:results_finetuned} report the experimental results in terms of BERTScore, presenting the performance of baseline models and their fine-tuned counterparts, respectively, in comparison with the proposed framework.

In Table~\ref{tab:results_baseline}, the baseline models exhibit varying levels of performance, with Qwen3-0.6B achieving the highest score of 82.00\%, while Aragpt2-Base records the lowest performance at 72.51\%. The proposed framework significantly outperforms all baseline models, achieving 86.71\%, which indicates a substantial improvement over individual model predictions. This result demonstrates the effectiveness of the proposed approach even without fine-tuning, highlighting the benefit of leveraging multiple models with a relevance-based selection mechanism.

\begin{table}[h]
\centering
\renewcommand{\arraystretch}{1.2}
\caption{Performance comparison of baseline models and \textbf{(Ours)}}
\label{tab:results_baseline}
\begin{tabular}{lc}
\hline
\textbf{Model} & \textbf{BERTScore} \\
\hline

Bloomz-560M & 78.72\% \\
Qwen3-0.6B & 82.00\% \\
Aragpt2-Medium & 79.69\% \\
Aragpt2-Base & 72.51\% \\
Dialect-ar-gpt-2021 & 81.61\% \\

\textbf{(Ours)} & \textbf{86.71\%} \\

\hline
\end{tabular}
\end{table}

Table~\ref{tab:results_finetuned} shows that fine-tuning consistently improves the performance of all models. For example, Aragpt2-Base increases from 72.51\% to 85.51\%, and Dialect-ar-gpt-2021 improves from 81.61\% to 86.74\%. Despite these gains, performance varies across models, suggesting that individual fine-tuned models still struggle to produce optimal responses for all queries consistently. In contrast, the proposed framework achieves the highest performance of 90.30\%, outperforming all fine-tuned models by a clear margin.

\begin{table}[h]
\centering
\renewcommand{\arraystretch}{1.2}
\caption{Performance comparison of fine-tuned models and \textbf{(Ours)}}
\label{tab:results_finetuned}
\begin{tabular}{lc}
\hline
\textbf{Model} & \textbf{BERTScore} \\
\hline

Bloomz-560M & 82.81\% \\
Qwen3-0.6B & 84.22\% \\
Aragpt2-Medium & 85.45\% \\
Aragpt2-Base & 85.51\% \\
Dialect-ar-gpt-2021 & 86.74\% \\

\textbf{(Ours)} & \textbf{90.30\%} \\

\hline
\end{tabular}
\end{table}

The observed improvements can be attributed to the integration of curriculum-based training and relevance-driven response selection. While fine-tuning enhances each model’s ability to generate more accurate responses, the proposed selection mechanism further refines the output by choosing the most semantically relevant response among multiple candidates. This combination reduces the impact of individual model weaknesses and improves overall robustness.

Overall, as illustrated in Figure \ref{fig:Results_chart} the results confirm that the proposed framework not only benefits from curriculum learning during training but also effectively utilizes multi-model diversity at inference time, leading to more accurate and reliable medical text generation compared to both baseline and fine-tuned approaches.

\section{Conclusion}

This paper presented a severity-aware multi-model framework for medical text generation that integrates severity-based training with relevance-driven response selection. The proposed approach addressed key limitations of existing methods, which often rely on single-model generation and do not explicitly account for the progressive complexity of medical cases or variability in response quality.

By adopting a severity-based curriculum learning strategy, the models were trained to gradually capture medical knowledge from mild to critical cases, leading to improved generalization and stability. In addition, the use of multiple LLMs introduced diversity in generated responses, while the BERTScore-based selection mechanism ensured that the final output was the most semantically aligned with the patient query. 

Experimental results demonstrated that the proposed framework consistently outperforms both baseline and fine-tuned models, achieving higher accuracy and more reliable responses. These findings highlight the effectiveness of combining curriculum learning with multi-model response selection in medical text generation.

Future work will explore more advanced relevance scoring methods beyond BERTScore, as well as integrating domain-specific knowledge to enhance further the quality and safety of generated medical responses. In addition, future studies will investigate the use of manually validated severity labels, merge medical expert evaluations, and assess the framework with respect to hallucination detection, harmful medical advice, and emergency-response safety to provide a more comprehensive evaluation of clinical reliability.

\bibliographystyle{IEEEtran}
\bibliography{ref}

\end{document}